\pdfoutput=1

\documentclass[11pt]{article}

\usepackage[preprint]{acl}

\usepackage{times}
\usepackage{amsmath, amssymb, amsthm}
\usepackage{latexsym}
\usepackage{mathletters}
\usepackage{cuted}           
\usepackage{multirow}
\usepackage{booktabs}
\usepackage{enumitem}
\usepackage{inconsolata}

\usepackage[utf8]{inputenc}
\usepackage[ruled,vlined]{algorithm2e}

\usepackage{graphicx}
\usepackage{float}
\usepackage{subcaption}
\usepackage{microtype}

\usepackage{algorithmic}

\title{Faster Machine Translation Ensembling with Reinforcement Learning and Competitive Correction}

\author{Kritarth Prasad, Mohammadi Zaki, Pratik Singh \and Pankaj Wasnik \\
Media Analysis Group, Sony Research India, Bangalore\\
\{kritarth.prasad, mohammadi.zaki, pratik.singh, pankaj.wasnik\}@sony.com}

\begin{document}
\maketitle
\begin{abstract}
Ensembling neural machine translation (NMT) models to produce higher-quality translations than the $L$ individual models has been extensively studied. Recent methods typically employ a candidate selection block (CSB) and an encoder-decoder fusion block (FB), requiring inference across \textit{all} candidate models, leading to significant computational overhead, generally $\Omega(L)$. This paper introduces \textbf{SmartGen}, a reinforcement learning (RL)-based strategy that improves the CSB by selecting a small, fixed number of candidates and identifying optimal groups to pass to the fusion block for each input sentence. Furthermore, previously, the CSB and FB were trained independently, leading to suboptimal NMT performance. Our DQN-based \textbf{SmartGen} addresses this by using feedback from the FB block as a reward during training. We also resolve a key issue in earlier methods, where candidates were passed to the FB without modification, by introducing a Competitive Correction Block (CCB). Finally, we validate our approach with extensive experiments on English-Hindi translation tasks in both directions.

\end{abstract}

\section{Introduction}

Recent advancements in machine translation have led to the availability of various open-source neural machine translation (NMT) models and general-purpose Large Language Models (LLMs) capable of translation tasks. However, the performance of these models depends on the model size and the quality of the pre-training data. As a result, ensembling methods have become essential for achieving state-of-the-art performance when low-parameter models are the only viable option due to resource constraints.

\begin{figure}[!t]
    \centering
    \includegraphics[width=0.49\textwidth]{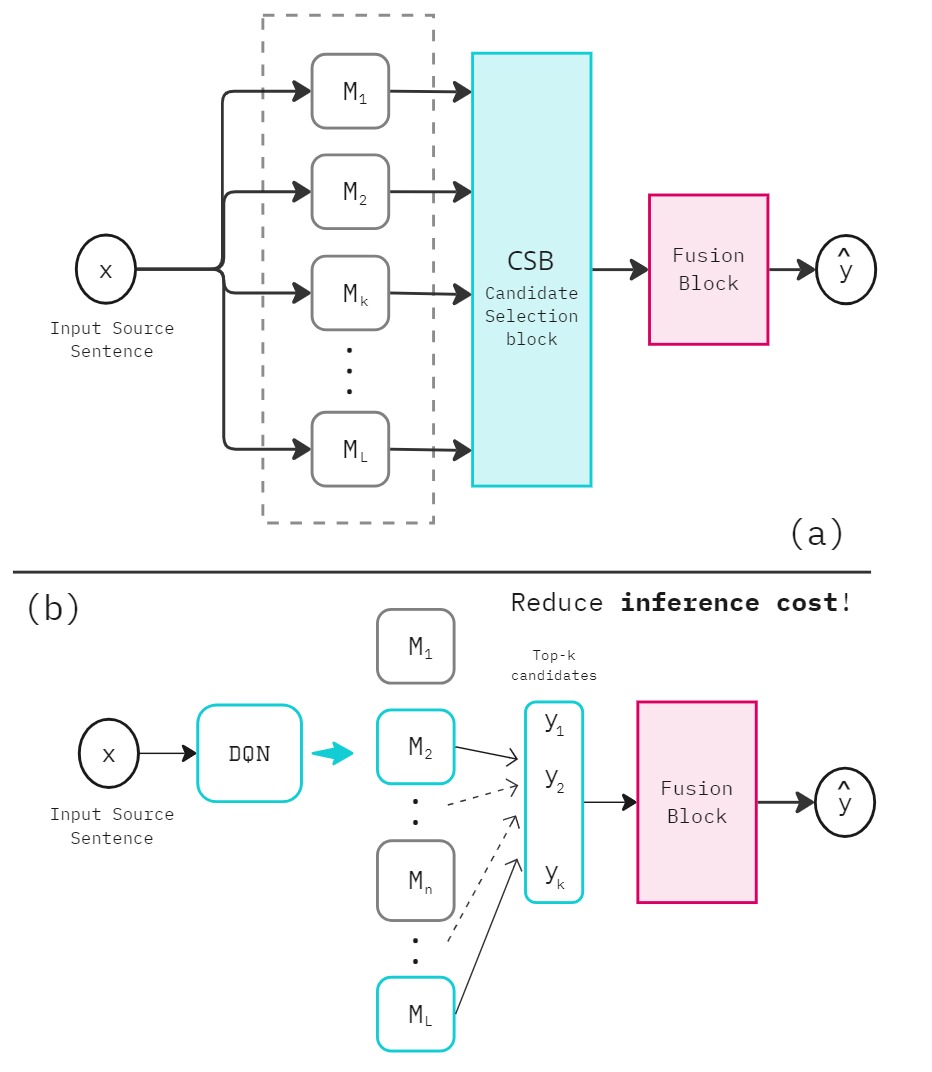}
    \caption{(a)The general strategy of ensembling MT systems, comprising of a candidate selection block followed by a fusion block, (b) Modified approach in this work.}
    \label{fig:general-ensemble-block}
\end{figure}

While traditional ensembling methods often rely on static weight sharing between models, these methods are now replaced by more dynamic approaches that adjust the weights of the component models depending on the specific source sentence and the quality of each candidate translation produced. A common approach to ensemble LLMs/MT models is depicted in Figure~\ref{fig:general-ensemble-block}; it comprises a candidate selection block (CSB) that selects a subset of candidate translations, potentially followed by a fusion block (FB), which is commonly an encoder-decoder network that combines the input to produce a (usually) better output translation.

These approaches, however, require the source sentence $\mathbf{x}$ to be translated by each of the $L$ candidate models in the first stage, which the CSB then takes as input. This significantly increases the time complexity of the ensembling process, making it a major drawback, particularly when the inference cost of LLMs is a fundamental challenge \cite{llm-inference-1, llm-inference-2, llm-inference-3}. Additionally, our extensive experiments reveal that current CSB strategies are suboptimal in selecting the best candidates to pass to the FB (discussed in detail in Sec.\ref{sec:DQN Motive}). This motivates us to address the need for an ensembling method that improves the quality of translations and deals with the high computational cost associated with traditional approaches. Specifically, we aim to design a strategy that delivers superior performance compared to state-of-the-art methods while reducing the inference time.


Thus, in this paper, we frame the candidate selection task as a reinforcement learning problem and employ a Deep Q-Network (DQN) block that efficiently identifies the group of optimal candidate translations (actions) for every source input sentence (state). Further, through experimental analysis, we identify that the overall performance of recent ensembling approaches is often limited by the worst-performing candidate in the system. To address this, we introduce a strategic candidate improvement method, which significantly enhances the performance of the fusion block.

Furthermore, the recent work of \citet{lu2023routingexpertefficientrewardguided} addresses the inference cost issue in current ensembling methods. However, their approach is a \textit{selection-based} ensembling, while we focus on \textit{generation-based} ensembling with the anticipation that the selected candidates can be further modified to produce potentially better translations. Our key contributions can be summarized as follows:
\begin{itemize}[leftmargin=*]
    \item We introduce \textbf{SmartGen}, a novel strategy that leverages DQN to dynamically select a subset of candidate MT models based on the source sentence, which are then fused in the next step. This approach significantly reduces the inference time of the ensembling method by only inferring from a subset of MT models (Figure~\ref{fig:cost-plot}).
    \item To enhance translation quality further, we propose a correction strategy (Sec.~\ref{subsec:CCB_method}) that selectively improves candidates using rejected ones. This combined methodology, referred to as \textbf{SmartGen++}, achieves significant performance gains with a modest increase in inference cost.
    \item We demonstrate the effectiveness of the proposed strategy through extensive experiments on English-to-Hindi and Hindi-to-English translation tasks, using four publicly available datasets and a private datasets, evaluated across multiple metrics.
\end{itemize}


\section{Background and Motivation}
\subsection{Previous Work}

{\bfseries Ensemble Learning and Ranking: } Ensemble learning is a widely used technique that combines the strengths of individual models to produce improved outputs \cite{ensemble_survey_1, ensemble_survey_2}. With the recent advancements in large language models, ensemble learning has become instrumental in employing multiple experts to tackle complex tasks based on their respective expertise \cite{multi-agent_1}. Studies have shown that ensembling can be achieved through model weights \cite{wan2024knowledgefusionlargelanguage, goddard2024arceesmergekittoolkitmerging} or by combining model outputs. In this context, the recent mixture of experts (MoE) approach has emerged as an effective technique for significantly boosting performance by merging outputs from multiple expert models \cite{MOE}.

Recent research introduces various selection methods prior to the fusion step, selecting only the best candidates from the experts for fusion. Significant work has been carried out in the field of summarization using different techniques such as training reranking models based on metrics \cite{summaranker}, employing contrastive learning for effective ranking \cite{simcls}, and using pair-wise ranking to compare candidates \cite{llm-blender}. However, there has been limited exploration of such methods in the task of machine translation \cite{hoang2024ontheflyfusionlargelanguage}. Moreover, existing work does not establish a clear connection between the top-ranked candidates and the final fusion output, as discussed in detail in Sec.~\ref{sec:DQN Motive}.

\noindent{\bfseries LLM Refinement: } As a result of continuous advancements in LLMs, their reasoning capabilities have significantly improved. This has prompted studies on enhancing performance through various strategic self-correction methods \cite{survey_selfcorrect}. Previous research has investigated self-scoring and self-correction techniques across various domains \cite{dynamic_llms, selfrefineselfiter}, along with the development of automated feedback systems \cite{instructscore}. In the context of NMT systems, common approaches include improvements through several iterations \cite{iterativeNMT} and post-editing strategies for correcting translations \cite{gpt_postedit}. However, self-improvements in selected candidates in ensembling are yet to be explored, as discussed in detail in \ref{subsec:CCB_method}.

\noindent{\bfseries RL for Ensembling and Model Selection:} 
Several notable works have employed reinforcement learning (RL), particularly multi-armed bandits, for model selection across various domains. Applications include load forecasting \cite{load-fore}, wind-speed prediction \cite{wind-speed}, airline pricing \cite{airline}, and the development of online and scalable model selection strategies \cite{mab-ms}, among others.

\subsection{Motivation}
\noindent{\bfseries Motivation for DQN Block:} \label{sec:DQN Motive} As discussed, the tried and tested approach to combine an ensemble of candidate translations to produce higher quality translations is by first comparing the candidate translations and scoring them, before passing the top $K$ candidates to an encoder-decoder block for fusion. However, this approach to handling output translations $(y_1, y_2, \ldots, y_L)$ is suboptimal, as the selection module is trained independently of the fusion block \cite{hoang-etal-2024-fly}. Specifically, the chosen candidates may perform very well individually, however, they may not be the best choice for the fusion block to produce the best results effectively.

\begin{figure*}[t!]
     \centering
    \begin{subfigure}[b]{.5\textwidth}
         \centering
         \includegraphics[width=7.5cm, height=4cm]{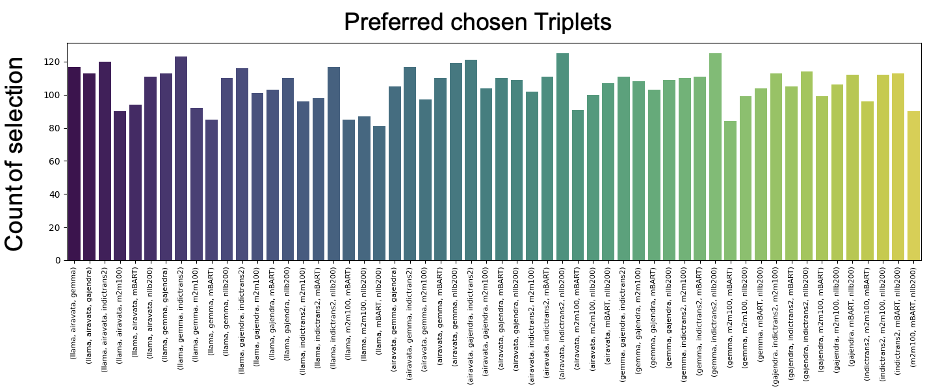}
         \caption{}
         \label{fig:labelled_our_Approach}
     \end{subfigure}%
     \hfill
     \begin{subfigure}[b]{.5\textwidth}
         \centering
         \includegraphics[width=7.5cm, height=4cm]{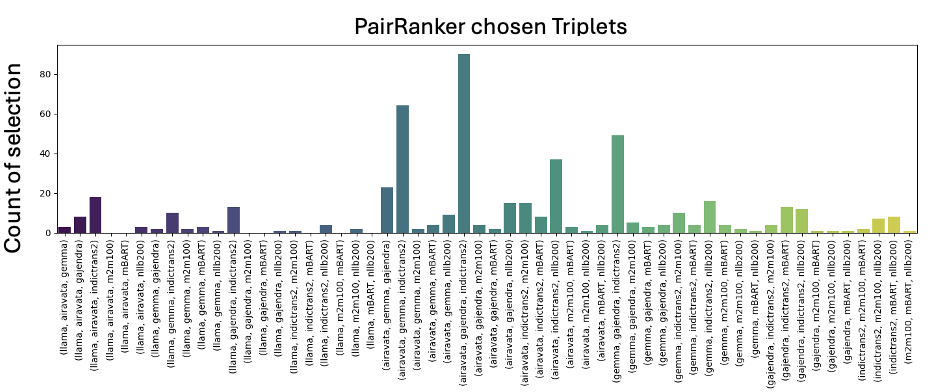}
         \caption{}
         \label{fig:errorbar}
     \end{subfigure}
     \begin{subfigure}[b]{.5\textwidth}
         \centering
         \includegraphics[width=7.5cm, height=4cm]{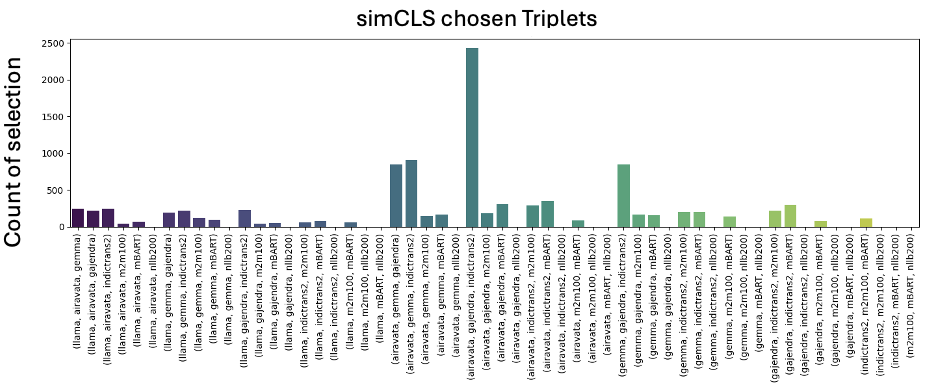}
         \caption{}
         \label{fig:labelled_our_Approach}
     \end{subfigure}%
     \hfill
     \begin{subfigure}[b]{.5\textwidth}
         \centering
         \includegraphics[width=7.5cm, height=4cm]{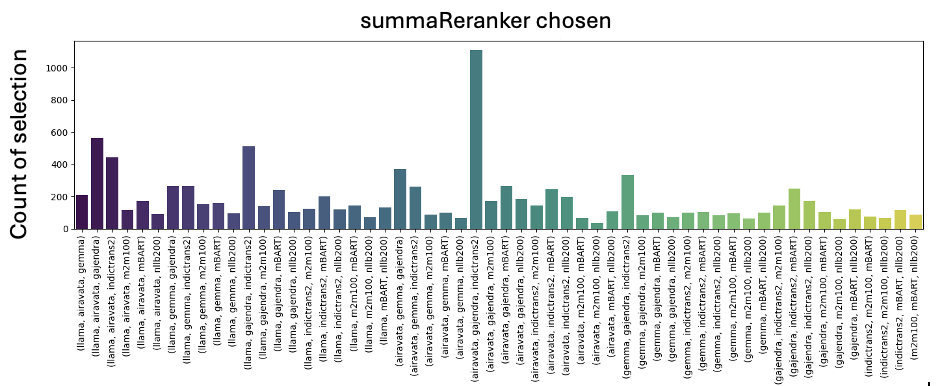}
         \caption{}
         \label{fig:labelled_our_Approach}
     \end{subfigure}%
     \hfill
        \caption{A comparative analysis of the distribution of the number of times candidate triplets are chosen in a) brute-force strategy,  b) Pair-Ranker \cite{llm-blender}, (c) SimCLS \cite{simcls}, and (d) SummaRanker \cite{summaranker} }
        \label{fig:why_dqn}
        
\end{figure*}

To test our assumption, we conducted an experiment using a pool of 8 MT systems for English-to-Hindi translation. Then for every sentence in a chosen test set, we observe by brute force the optimal triplet $(y_{n_1},y_{n_2},y_{n_3})$ of candidate translations (we choose $K=3$ here, in line with the choice in \citet{llm-blender}) which leads to achieving the best performance (in terms of BLEU score) when passed through the fusion block. 
In Figure~\ref{fig:why_dqn}, we compare the number of times various SOTA ensembling methods select a specific triplet with the brute-force strategy described earlier, which determines the optimal triplet every time. The plot demonstrates that for different sentences in the test sets, different triplets yield optimal performance, with no clear preference for any particular subset across all source sentences. In contrast, SOTA algorithms tend to favor certain triplets, leading to a gap between the achieved and the achievable performance.

We instead opt to choose a dynamic strategy, in that it explores the available action space for possibly better candidate groups. We, therefore, choose a DQN for action selection strategy and train it with the help of the reward obtained from the \emph{final} translation obtained from the fusion block, instead of treating it independently.

\noindent{\bfseries Motivation for the Competitive Correction Block:} \label{sec:CCB} Another crucial observation that we make is that the performance of the fusion block is limited by and degrades with the `worst' input candidate translation. To verify this, we perform an experiment where 1) we pass the `reference` (actual translation) sentence to the fusion block repeatedly $K$ times, and compare it with 2) reference translation + top $(K-1)$ translations from the candidate pool. We report the result in Table~\ref{tab:oracle vs noise}. This simple experiment demonstrates that even though the selection block successfully manages to pick the best candidates from the pool, the performance of the final translation severely degrades (BLEU decreased by more than 17 points) depending on the `other' chosen translations.

\begin{table}[!ht]
\small
\centering
\begin{tabular}{|c|c|}
\hline
\textbf{Candidates}                     & \textbf{BLEU} \\ \hline
Reference  $\times K$                           & 85.83         \\ \hline
Reference + Top $K-1$ BLEU  candidates & 68.38         \\ \hline
\end{tabular}
\caption{A comparison of the performance of the reference sentence repeated $K$ times vs. reference appended with $K-1$ other candidates. The task is carried out for English to Hindi on a subset of human translated test set.}
\label{tab:oracle vs noise}
\end{table}

This limitation restricts all methods following the usual process of ensembling (Figure~\ref{fig:general-ensemble-block}), ie., select and fuse. To mitigate this issue, we introduce a candidate improvement strategy, which we explain in detail in Sec.~\ref{subsec:CCB_method}, which aims to identify and subsequently improve the weak candidate translations from the set of selected candidates before passing to the fusion block.

\section{Proposed Methodology}\label{sec:methodology}

In this section, we first present the problem statement and define the necessary notations, followed by the details of the proposed methodology. \\
\noindent{\bfseries Problem Statement:} Given a set of $L$ machine translation (MT) systems (e.g., NMT systems, LLMs), denoted as ${M_1, M_2, \ldots, M_L}$, the objective is to translate a source sentence $\mathbf{x}$ into a target language such that the output sentence $\hat{y}$ has higher quality than any of the individual candidate translations $(y_1, y_2, \ldots, y_L)$, while minimizing the computational overhead during inference.

\begin{figure*}[ht]
    \centering
    \includegraphics[width=\textwidth]{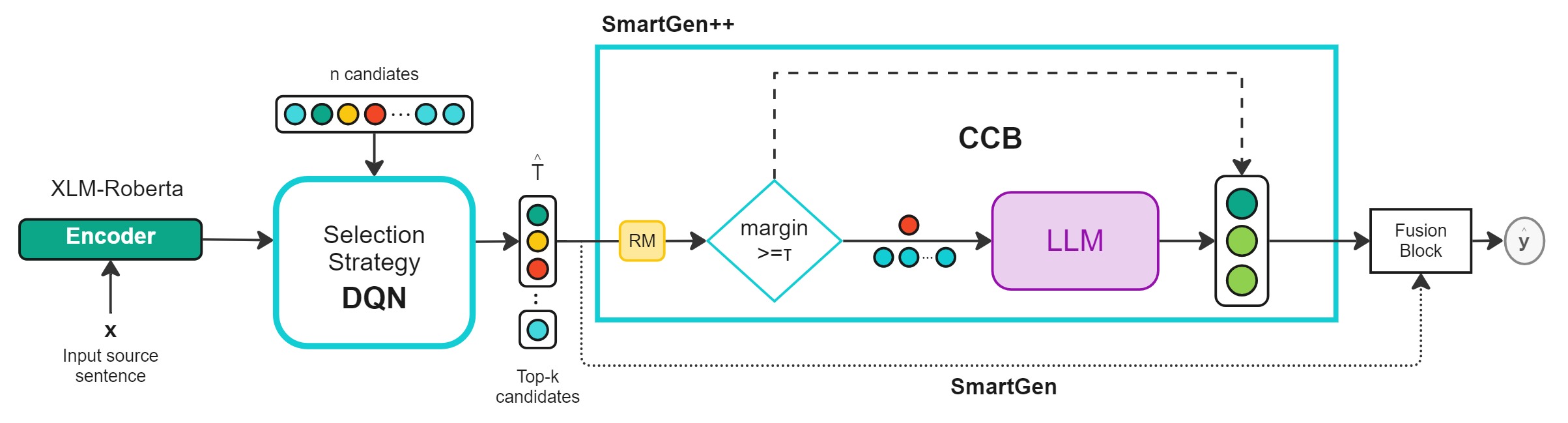} 
    \caption{Detailed block diagram of our proposed methodology. Here DQN+FB represents SmartGen, and DQN+CCB+FB represents SmartGen++ as explained in text.}
    \label{fig:main-block}
\end{figure*}

The proposed methodology proceeds through three main phases: 1) candidate translation selection via the DQN network, 2) improvement of the selected translations with the Competitive Candidate Improvement (CCI) block, and 3) Fusion of the selected candidates with standard Encoder-Decoder \cite{llm-blender} block. A block diagram of SmartGen(++) is shown in Figure~\ref{fig:main-block}. We present details of these modules in the subsequent sections. 

\subsection{Candidate Translation Selection using DQN Block}\label{subsec:DQN}
Here, we employ a Deep Q-Network (DQN) \cite{dqn} to select the group of translations that we pass to the fusion block. Formally, we model the action selection process as a discounted Markov Decision Process (MDP) $\cM(\cS,\cA,\cP,r,\gamma)$, where $\cS$ represents the state space, consisting of all possible embedding vectors of the input sentence $\mathbf{x}$. The action space $\cA$ is defined as the set ${1, 2, \ldots, L}$. We select the candidates $\hat{T}:={y_{n_1}, y_{n_2}, \ldots, y_{n_K}}$ to pass to the fusion block by choosing $\arg , TopK_{a} , Q(s,a)$. Further, $\gamma$ is a discount factor whose value is in the range $[0, 1)$. The $\cP$ is a probability transition kernel, $\cP:\cS\times \cA\to \Delta_{\cS}$, which gives a distribution over the next state, given the current state and the action.

        We use a standard experience replay buffer \cite{dqn} to store and resample tuples $\langle s, a, s', r \rangle$ for training the DQN network, where $r:\cS \times \cA \to \mathbb{R}$ is a function that assigns a scalar reward based on the action taken in a given state. To train the DQN, we use the sacreBLEU \cite{bleu} score (normalized between [0,1]) of the final output translation $\hat{y}$ generated by the fusion block.

\subsection{Competitive Correction Block (CCB)}\label{subsec:CCB_method}

Building on the motivation from Section \ref{sec:CCB}, we introduce the Competitive Correction Block (CCB), designed to selectively enhance the quality of input translations passed to the fusion block. This strategy can be applied to other ensemble approaches that rely on a select-and-fuse mechanism, making it of independent interest for improving final translation quality. The proposed CCB block consists of two key components: 1) the reward model and 2) the correction block.

\noindent{\bfseries{Reward Model} (RM):} To evaluate the quality of selected candidates, we develop a Reward Model (RM) specifically designed to align Large Language Models (LLMs) with human preferences, providing a nuanced reward signal. Traditional reward modeling frameworks \cite{stiennon2022learningsummarizehumanfeedback,ouyang2022traininglanguagemodelsfollow} face limitations due to their bias towards a \textit{single} preferred response and their inability to capture contextual relationships among \textit{multiple} similar candidates— a critical factor in translation tasks where multiple candidates may be very similar. To address this, we introduce a more comprehensive training methodology that considers a \textit{set} of both preferred and rejected responses for each sample. Specifically, for each input sample $\mathbf{x}$, we select \textit{four} preferred responses, denoted by $\cP$ (including human preference and the top-3 BLEU candidates), with the remaining candidates designated as rejected, denoted by $\cR$. This approach ensures a diverse, high-quality ensemble of responses that reflects the complexity of human preferences and the contextual relationships between candidates and reward scores. Thus, we can write modified RM loss as:
\begin{equation}
\begin{aligned}
\mathcal{L}_{\text{RM}}(\theta) = \ & -\mathbb{E}_{\mathbf{x} \sim \cD, \, y_p \in \mathcal{P}, \, y_r \in \mathcal{R}} \bigg[ \\
& \log \sigma \big( r_\theta(\mathbf{x}, y_p) - r_\theta(\mathbf{x}, y_r) \big) \bigg]
\end{aligned}
\end{equation}
where $r_\theta(\mathbf{x}, y)$ is the scalar output of the reward model for input sample $\textbf{x}$ and target candidates $y$ with parameters $\theta$, $y_p$ and $y_r$ represents preferred and rejected candidates respectively and $\cD$ is the dataset used for training. The RM block is used to evaluate the quality of the selected candidates $\hat{T}$, and depending on a user-specified `margin', candidates are passed to the correction block, described next.

\noindent{\bfseries Correction Block:}
The functioning of the Correction Block (CB) is outlined in Alg.~\ref{alg:CCB_algo}. Specifically, based on the reward margin between the selected candidates, we aim to enhance their quality using an LLM $\cG$ and the set of rejected candidates from the DQN. We anticipate that by providing candidate translations and their respective rewards to $\cG$, the model can generate a new candidate, $\hat{I}$, that surpasses the current reward of $I$ and the rewards of the candidates in $[L]\backslash\hat{T}$. The exact prompt used for this process is detailed in the in Figure \ref{fig:ccbprompt}
 
    
\begin{algorithm}[h]\label{alg:CCB_algo}
    \caption{Competitive Correction Block (CCB)}\label{alg:CCB_algo}
    \textbf{Terminologies:} DQN ranked output (sorted based on rewards): $\hat{T}$,  Rewards of $\hat{T}$: ${R}_{\hat{T}}$ , Rejected Candidates by DQN: $[L]\backslash\hat{T}$, threshold: $\tau$, Enhancer LLM: $\cG$, Candidates of $[L]\backslash\hat{T}: \{c_1, c_2, c_3, ..., c_{L-K}\}$, Reward of $L\backslash\hat{T}: \{b_1, b_2, ..., b_{L-K}\}$ \\ 
    \textbf{Input:}  $\cG$, $\tau$, $\hat{T}$, ${R}_{\hat{T}}$,  $[L]\backslash\hat{T}$ \\
    ${R}_{\hat{T}} \gets \{r_1, r_2,.., r_K\}$ \\
    $margin \gets \{r_1, r_1-r_2, ..., r_{K-1}-r_K\}$ \\
    
    \For{$m$ in $2 : (K)$}{
        \If{$margin[m] \geq \tau$}{
            $I \gets \hat{T}[m]$ \\
            $R_I \gets {R}_{\hat{T}}[m]$ \\
            $\hat{I} \gets \cG((I, R_I),\{(c_{1:L-K}, b_{1:L-K})\}$ \\
            $\hat{T}\gets \hat{T}\backslash I \cup \hat{I}$
        }    
    }
   \textbf{Output: } $\hat{T}$
\end{algorithm}

\section{Experiments}
In this section, we present the experiments conducted to evaluate the effectiveness of our proposed methodology, detailing the datasets, baselines, setup and key results.

\subsection{Datasets} 
For our experiments, we focus primarily on English-to-Hindi and Hindi-to-English translation tasks, using a diverse set of open-source and private datasets for training and evaluation. Detailed dataset statistics can be found in the Appendix in Table~\ref{tab:data_stats}. The reward model was trained on a high-quality, privately collected dataset, and evaluations were conducted on benchmark datasets such as Flores \cite{flores}, WMT-14 \cite{wmt14}, IN22-Gen, and IN22-Conv \cite{bpcc} along with entertainment-domain private data.


\subsection{Baselines} 
For candidate translation generation, we utilize a combination of large language models (LLMs) such as Llama-3-8B \cite{llama3}, Gemma-2-9B \cite{gemma_2024}, Airavata \cite{gala2024airavata}, Gajendra, and state-of-the-art machine translation models including IndicTrans2 \cite{gala2023indictrans}, NLLB200 \cite{nllbteam2022languageleftbehindscaling}, m2m100 \cite{m2m100}, and mBART \cite{mbart}. For reward model training, we leverage the pre-trained Bloom-1b7 \cite{workshop2023bloom176bparameteropenaccessmultilingual}. The fusion model is based on mT5-large \cite{fuser}, and mdeberta-v3-base \cite{he2021deberta} is used for all baseline ranking systems. For encoding candidates during DQN state generation, we employ XLM-RoBERTa \cite{conneau2020unsupervisedcrosslingualrepresentationlearning}. For GPT scoring we have GPT-4o \cite{openai2024gpt4technicalreport} 

\subsection{Evaluations Metrics}
We use SacreBLEU \cite{bleu} as the primary evaluation metric for our experiments. Additionally, we evaluate our models using other popular machine translation metrics such as chrF++ \cite{chrf}, WMT22-Comet-DA \cite{rei-etal-2020-unbabels} (referred to as Comet). 

 \subsection{Training Details}\label{subsec:training details}
For training of the baseline rankers (SimCLS \cite{simcls}, SummaReranker \cite{summaranker}, PairRanker \cite{llm-blender}), we utilized the default setting from \citet{llm-blender}. 
For DQN training, we implemented a ResNet-based DQN \cite{dqn, he2015deepresiduallearningimage} as the base model. The architecture begins with a linear layer that maps the input state dimension of 768 to a hidden dimension of 256, followed by three residual blocks. Each residual block contains two fully connected layers with ReLU activations and skip connections to improve gradient flow and model performance. The network concludes with an output linear layer that generates Q-values for the $L$ possible actions. We trained the DQN using a 10\% uniformly sampled subset of the dataset. The model converged after a few episodes, each consisting of 1,000 steps. The DQN training was performed on 1 x NVIDIA H100 80GB GPU, with convergence time varying between 24 to 48 hours.
For training the fusion block, we employed an mT5-large model, instructed with the prompt: ``\textit{Translate the source text into the target language.}".
The fusion block was trained using a configuration of 3 x NVIDIA H100 80GB GPUs. The main hyperparameters used in the experiments are shown in Table~\ref{tab:dqn-hyperparameters}.

\begin{table*}[ht]
    \centering
    \resizebox{\textwidth}{!}{  
        \begin{tabular}{c c @{\hspace{0.5cm}}|ccc|ccc|ccc|ccc|ccc}
            \toprule
            \textbf{Category} & \textbf{Methods} & 
            \multicolumn{3}{c|}{\textbf{Private data}} & 
            \multicolumn{3}{c|}{\textbf{IN22-Conv}} & 
            \multicolumn{3}{c|}{\textbf{IN22-Gen}} & 
            \multicolumn{3}{c|}{\textbf{FLORES}} & 
            \multicolumn{3}{c}{\textbf{WMT-14}} \\
            \cmidrule(lr){3-5} \cmidrule(lr){6-8} \cmidrule(lr){9-11} \cmidrule(lr){12-14} \cmidrule(lr){15-17}
                            &                    & 
            BLEU & COMET & chrF++ & 
            BLEU & COMET & chrF++ & 
            BLEU & COMET & chrF++ & 
            BLEU & COMET & chrF++ & 
            BLEU & COMET & chrF++ \\
            \midrule
            \multirow{8}{*}{\textbf{LLMs / NMTs}} 
                & LLaMA-3-8B & 
                    44.50 & 83.45 & 56.38 & 
                    34.67 & 86.81 & 50.57 & 
                    26.44 & 82.39 & 47.10 & 
                    32.31 & 85.45 & 53.45 & 
                    27.16 & 82.83 & 47.43 \\
                & Gemma-2-9B & 
                    58.08 & 88.89 & 68.73 & 
                    43.16 & 90.41 & 58.42 & 
                    39.83 & 87.94 & 61.28 & 
                    43.18 & 88.92 & 63.22 & 
                    37.60 & 87.54 & 58.73 \\
                & Gajendra & 
                    31.51 & 73.44 & 39.61 & 
                    24.72 & 77.99 & 38.00 & 
                    23.25 & 78.90 & 43.10 & 
                    25.79 & 79.24 & 44.97 & 
                    21.07 & 74.89 & 37.27 \\
                & Airavata & 
                    45.53 & 81.66 & 54.32 & 
                    37.03 & 85.30 & 49.88 & 
                    30.42 & 82.39 & 50.24 & 
                    36.05 & 83.74 & 53.93 & 
                    30.71 & 81.24 & 47.80 \\
                & Indictrans2 & 
                    60.63 & 89.43 & 70.53 & 
                    46.48 & 90.69 & 60.04 & 
                    46.09 & 89.06 & 65.26 & 
                    51.60 & 90.39 & 68.12 & 
                    43.99 & 88.61 & 61.97 \\
                & m2m100 & 
                    42.15 & 81.77 & 54.42 & 
                    34.16 & 84.54 & 49.09 & 
                    29.47 & 82.84 & 52.35 & 
                    39.24 & 86.28 & 58.73 & 
                    31.53 & 82.26 & 51.59 \\
                & mBART & 
                    47.17 & 83.98 & 59.87 & 
                    37.19 & 86.81 & 52.52 & 
                    32.95 & 85.78 & 55.96 & 
                    37.90 & 86.65 & 58.42 & 
                    32.58 & 85.05 & 54.05 \\
                & nllb200 & 
                    50.22 & 86.52 & 61.62 & 
                    43.79 & 90.14 & 57.88 & 
                    40.41 & 88.05 & 61.02 & 
                    47.05 & 89.43 & 64.74 & 
                    40.95 & 87.85 & 59.55 \\
            \midrule
            \multirow{2}{*}{\textbf{Analysis}} 
                & Random & 
                    48.88 & 83.89 & 58.67 & 
                    38.90 & 86.73 & 52.70 & 
                    34.46 & 85.27 & 55.10 & 
                    39.93 & 86.49 & 58.42 & 
                    34.73 & 83.99 & 52.63 \\
                & Oracle (BLEU) & 
                    68.45 & 90.06 & 75.87 & 
                    54.85 & 91.38 & 66.38 & 
                    51.18 & 89.13 & 67.40 & 
                    56.78 & 90.31 & 71.02 & 
                    49.97 & 88.77 & 65.70 \\
            \midrule
            \multirow{4}{*}{\textbf{Rankers}} 
                & {simCLS} & 
                      56.03 & 87.21 & 65.29 & 
                      39.88 & 87.78 & 53.81 & 
                      39.55 & 87.14 & 59.05 & 
                      43.74 & 88.48 & 61.62 & 
                      39.41 & 86.63 & 57.18 \\
                & {summaReranker} & 
                      55.06 & 86.73 & 64.76 & 
                      38.92 & 87.76 & 54.02 & 
                      35.23 & 85.81 & 57.30 & 
                      40.88 & 87.64 & 60.79 & 
                      35.81 & 85.15 & 55.58 \\
                & {PairRanker} & 
                      58.14 & 88.43 & 67.49 & 
                      44.71 & 90.30 & 58.70 & 
                      41.80 & 88.22 & 62.03 & 
                      47.31 & 89.49 & 65.11 & 
                      41.17 & 87.96 & 60.27 \\
                & {DQN (proposed)} & 
                      61.75 & 89.59 & 70.75 & 
                      45.78 & 90.51 & 59.55 & 
                      43.26 & 88.67 & 63.01 & 
                      48.79 & 89.90 & 66.09 & 
                      42.47 & 88.40 & 60.81 \\
            \midrule \midrule
            \multirow{3}{*}{\textbf{Blenders}} 
                & {LLM-Blender} & 
                      60.13 & 89.43 & 70.01 & 
                      44.72 & 90.30 & 58.70 & 
                      41.76 & 88.21 & 61.90 & 
                      47.25 & 89.48 & 65.09 & 
                      41.12 & 87.95 & 60.26 \\
                & {SmartGen} & 
                      58.34 & 88.47 & 68.07 & 
                      44.59 & 89.85 & 58.35 & 
                      42.82 & 88.29 & 62.76 & 
                      48.50 & 89.71 & 65.72 & 
                      41.78 & 87.96 & 60.15 \\
                & {SmartGen++} & 
                      62.77 & 90.09 & 70.75 & 
                      45.78 & 90.51 & 59.56 & 
                      43.26 & 88.67 & 62.96 & 
                      48.73 & 89.89 & 66.07 & 
                      42.42 & 88.39 & 60.79 \\
            \bottomrule
        \end{tabular}
    }
    \caption{Performance Metrics of Various Models Across Different Datasets in Translation from Hindi to English}
    \label{tab:perfhin}
\end{table*}

\begin{table*}[ht]
    \centering
    \resizebox{\textwidth}{!}{  
        \begin{tabular}{c c @{\hspace{0.5cm}}|ccc|ccc|ccc|ccc|ccc}
            \toprule
            \textbf{Category} & \textbf{Methods} & 
            \multicolumn{3}{c|}{\textbf{Private data}} & 
            \multicolumn{3}{c|}{\textbf{IN22-Conv}} & 
            \multicolumn{3}{c|}{\textbf{IN22-Gen}} & 
            \multicolumn{3}{c|}{\textbf{FLORES}} & 
            \multicolumn{3}{c}{\textbf{WMT-14}} \\
            \cmidrule(lr){3-5} \cmidrule(lr){6-8} \cmidrule(lr){9-11} \cmidrule(lr){12-14} \cmidrule(lr){15-17}
                            &                    & 
            BLEU & COMET & chrF++ & 
            BLEU & COMET & chrF++ & 
            BLEU & COMET & chrF++ & 
            BLEU & COMET & chrF++ & 
            BLEU & COMET & chrF++ \\
            \midrule
            \multirow{8}{*}{\textbf{LLMs / NMTs}} 
                & {LLaMA-3-8B} & 
                  24.08 & 71.56 & 39.94 & 
                  18.71 & 76.05 & 35.93 & 
                  18.77 & 68.65 & 38.61 & 
                  20.63 & 69.61 & 39.53 & 
                  19.52 & 70.76 & 37.37 \\
                & {Gemma-2-9B} & 
                  33.26 & 79.09 & 49.34 & 
                  26.64 & 82.48 & 43.34 & 
                  25.39 & 70.92 & 45.66 & 
                  33.35 & 77.61 & 52.72 & 
                  27.44 & 77.53 & 45.70 \\
                & {Gajendra} & 
                  35.08 & 80.23 & 50.05 & 
                  30.81 & 83.86 & 46.57 & 
                  36.25 & 79.71 & 53.68 & 
                  36.22 & 80.25 & 55.39 & 
                  30.64 & 80.31 & 48.40 \\
                & {Airavata} & 
                  36.68 & 81.73 & 53.12 & 
                  30.72 & 84.23 & 46.74 & 
                  37.83 & 80.51 & 55.23 & 
                  38.12 & 81.32 & 57.63 & 
                  32.15 & 81.37 & 50.00 \\
                & {Indictrans2} & 
                  41.96 & 82.92 & 57.46 & 
                  32.96 & 85.10 & 48.93 & 
                  37.90 & 80.48 & 55.83 & 
                  40.01 & 81.59 & 59.63 & 
                  33.34 & 81.68 & 51.39 \\
                & {m2m100} & 
                  27.34 & 76.14 & 43.79 & 
                  23.75 & 80.57 & 40.29 & 
                  23.86 & 72.79 & 43.80 & 
                  31.81 & 75.92 & 50.90 & 
                  24.67 & 75.93 & 42.87 \\
                & {mBART} & 
                  25.84 & 76.00 & 43.36 & 
                  25.82 & 81.30 & 41.42 & 
                  25.10 & 76.53 & 45.94 & 
                  26.64 & 76.51 & 46.30 & 
                  24.94 & 76.91 & 42.81 \\
                & {nllb200} & 
                  30.60 & 74.63 & 44.60 & 
                  29.95 & 82.61 & 45.34 & 
                  16.11 & 66.73 & 36.03 & 
                  26.98 & 71.09 & 45.73 & 
                  22.29 & 72.85 & 39.88 \\
            \midrule
            \multirow{2}{*}{\textbf{Analysis}} 
                & {Random} & 
                  45.42 & 83.30 & 59.41 & 
                  32.22 & 84.60 & 48.07 & 
                  28.86 & 77.81 & 48.85 & 
                  34.33 & 80.08 & 53.75 & 
                  28.02 & 79.70 & 46.35 \\
                & {Oracle (BLEU)} & 
                  54.41 & 85.23 & 66.73 & 
                  77.31 & 93.18 & 85.97 & 
                  66.61 & 85.48 & 76.81 & 
                  70.31 & 87.77 & 83.66 & 
                  35.41 & 81.79 & 52.82 \\
            \midrule
            \multirow{4}{*}{\textbf{Rankers}} 
                & {simCLS} & 
                  41.01 & 82.07 & 54.79 & 
                  32.98 & 84.84 & 47.88 & 
                  30.47 & 75.95 & 50.00 & 
                  35.95 & 80.60 & 55.24 & 
                  30.11 & 80.58 & 48.10 \\
                & {summaReranker} & 
                  37.00 & 79.05 & 50.50 & 
                  30.12 & 82.29 & 44.91 & 
                  26.15 & 74.70 & 45.38 & 
                  30.28 & 76.59 & 49.40 & 
                  25.96 & 77.25 & 43.84 \\
                & {PairRanker} & 
                  42.93 & 82.98 & 57.36 & 
                  33.85 & 85.15 & 49.14 & 
                  32.12 & 78.62 & 52.29 & 
                  37.33 & 81.13 & 57.11 & 
                  30.78 & 81.14 & 47.42 \\
                & {DQN (proposed)} & 
                  42.70 & 82.65 & 57.03 & 
                  30.60 & 83.37 & 46.22 & 
                  33.89 & 77.67 & 51.48 & 
                  36.60 & 80.44 & 55.65 & 
                  30.71 & 79.42 & 47.91 \\
            \midrule \midrule
            \multirow{3}{*}{\textbf{Blenders}} 
                & {LLM-Blender} & 
                  49.11 & 84.52 & 62.50 & 
                  34.86 & 85.67 & 50.54 & 
                  30.75 & 78.69 & 50.83 & 
                  36.92 & 81.41 & 56.77 & 
                  29.54 & 81.02 & 48.49 \\
                & {SmartGen} & 
                  48.13 & 84.12 & 61.60 & 
                  33.71 & 85.29 & 49.97 & 
                  32.29 & 79.10 & 51.89 & 
                  37.33 & 81.16 & 56.77 & 
                  29.86 & 80.57 & 48.52 \\
                & {SmartGen++} & 
                  49.45 & 84.67 & 62.83 & 
                  34.68 & 85.73 & 50.66 & 
                  32.51 & 79.50 & 52.48 & 
                  37.93 & 81.52 & 57.52 & 
                  30.21 & 81.20 & 49.20 \\
            \bottomrule
        \end{tabular}
    }
    \caption{Performance Metrics of Various Models Across Different Datasets in Translation from English to Hindi}
    \label{tab:perfeng}
\end{table*}

 

\subsection{Results}
\begin{figure}[!t]
    \centering
    \includegraphics[width=0.5\textwidth]{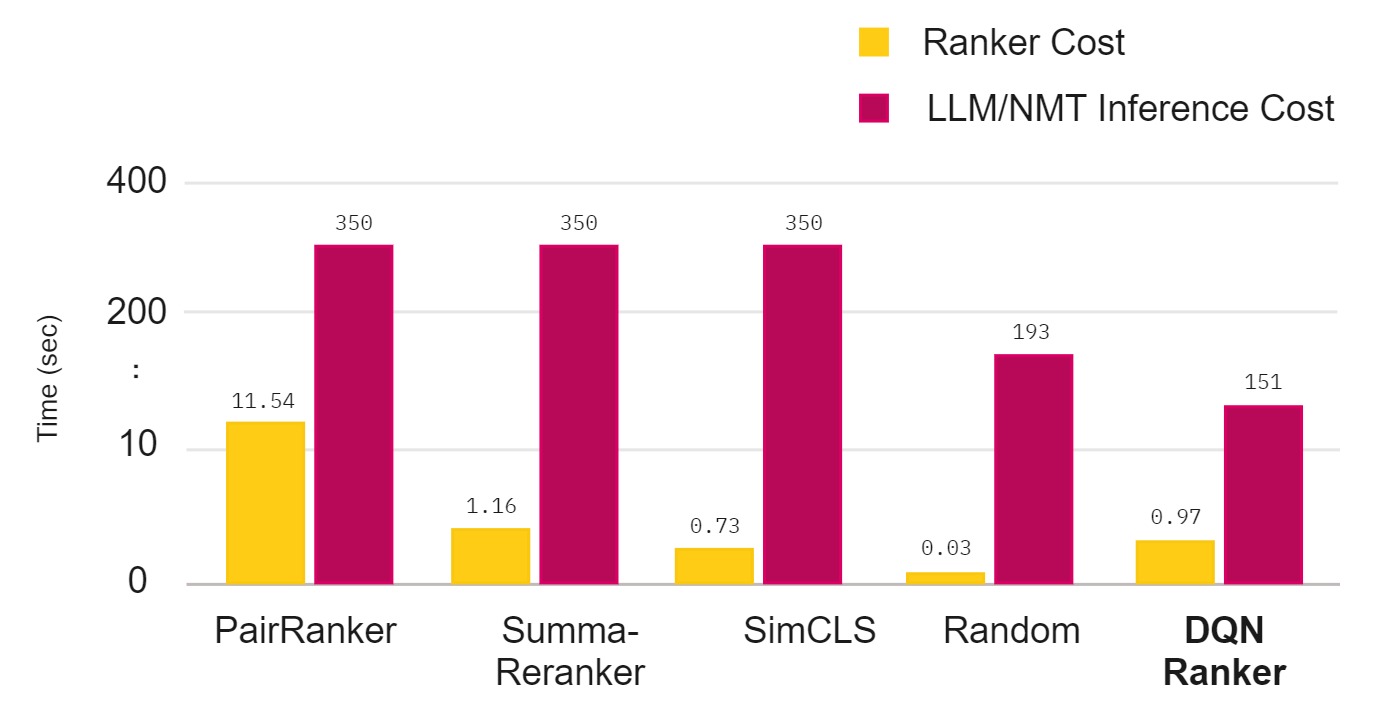}
    \caption{Demonstration of Inference Time Complexity/Cost among Rankers with their Ensemble LLMs Consumption; Reflecting DQN lowest Inference cost and better than all the baseline on 100 Uniformly Sampled test-set.}
    \label{fig:cost-plot}
\end{figure}

In our study, we evaluate the performance of our proposed methodologies, namely SmartGen and SmartGen++, against a spectrum of ReRanking baselines and individual expert systems. We also show results in the case where we directly output the best candidate selected by the DQN, which we call DQN(proposed) in Table~\ref{tab:perfeng} and \ref{tab:perfhin}. This evaluation is conducted across a range of automatic metrics including BLEU, Comet, and chrF++. Our analysis encompasses bidirectional translation tasks, specifically in English and Hindi, whose outcomes are detailed in Table \ref{tab:perfeng} and \ref{tab:perfhin}.\\

\noindent\textbf{SmartGen is Fast: }
In ranking-based ensembling systems, two types of costs are associated, namely, 1) \textbf{Ranking cost}, which refers to the inference time taken by the ranker, and 2) NMT systems \textbf{inference cost}, which is the total time taken by Candidate NMT systems to generate the translations.
From Figure~\ref{fig:cost-plot}, we observe that:
1) For Ranker Cost, Random Selection of Candidates is the fastest, followed by SimCLS which makes it fastest in baselines, and then SmartGen, which demonstrates better computational time compared to other baselines such as SummaReranker and LLM-Blender. LLM-Blender is the slowest, taking almost 10 times longer than any other baselines.
2) For Inference time of NMT systems, SmartGen takes almost 2.31 times more faster than any of the Rankers and is also 1.26 times faster than selecting random NMT systems, making it the fastest among all the systems.

\noindent\textbf{DQN Ranker is the best Trade-off: } 
We judged our approach against baselines ranking methods for the bidirectional translation task. 

\noindent\textit{\bfseries Hindi to English:} From Table \ref{tab:perfhin} we infer that our approach outperforms best baselines i.e PairRanker by 4.48\% on BLEU, 0.60\% on Comet and by 2.10\% on chrF++, while against simCLS, we beat it by 11.40\% on BLEU, 2.24\% on comet and 7.83\% on chrF++, suggesting that our training objectives were helpful.

\begin{figure}
    \centering
    \includegraphics[width=.95\linewidth]{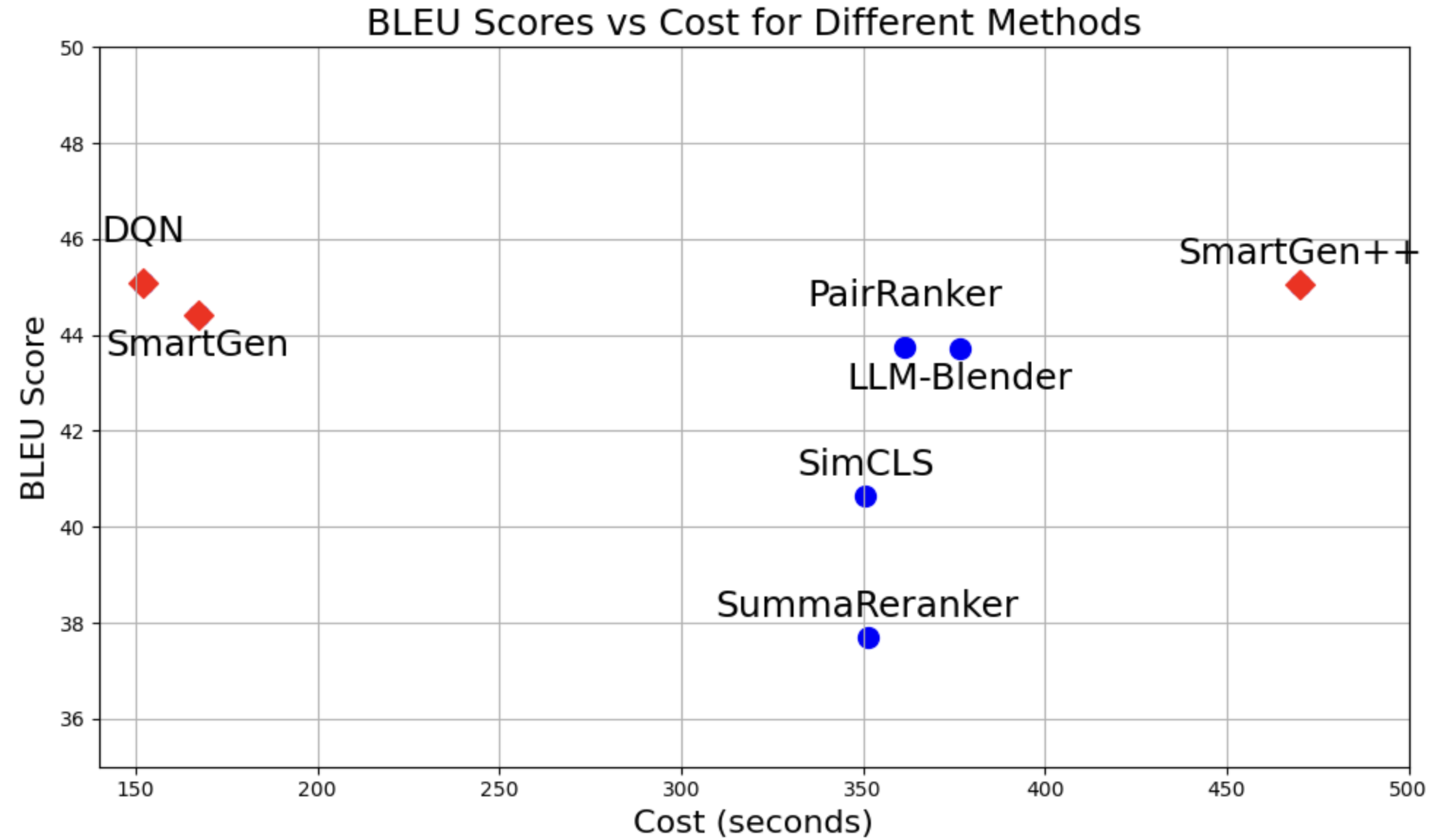}
    \caption{A scatter plot showing the tradeoff of various ensemble methods in terms of quality vs inference cost.}
    \label{fig:BLEUVSCOST}
\end{figure}

\noindent\textit{\bfseries English to Hindi}:  Furthermore, from Table \ref{tab:perfeng} we can infer that on average across datasets DQN Ranker have scores of 34.9 (BLEU), 80.71 (Comet) and 51.65 (chrF++), whereas best baseline PairRankers performance is 35.41 (BLEU), 80.80 (Comet) and 52.66 (chrF++). we can see that our DQN ranker scores are almost comparable with best ranker baseline and with very high computational efficiency which can be seen in the Figure \ref{fig:BLEUVSCOST}. The ranking methods outperformed random selection, except for summaReranker, which performed below the mark compared to baselines and even random selection.

\noindent\textbf{SmartGen is better: } We test SmartGen against LLM-Blender \cite{llm-blender} as it is the only baseline that uses Fusion Block after ranking. For English to Hindi direction SmartGen outperforms LLM-Blender very slightly on (0.22) BLEU on an average across all the test datasets. Moreover for Hindi to English we see SmartGen, outperforms LLM-Blender by 0.21 (BLEU) and is marginally better in terms of chrF++, on average across all testsets.

\subsection{Ablation Studies}

\begin{figure}[!ht]
    \centering
    \includegraphics[width=0.4\textwidth]{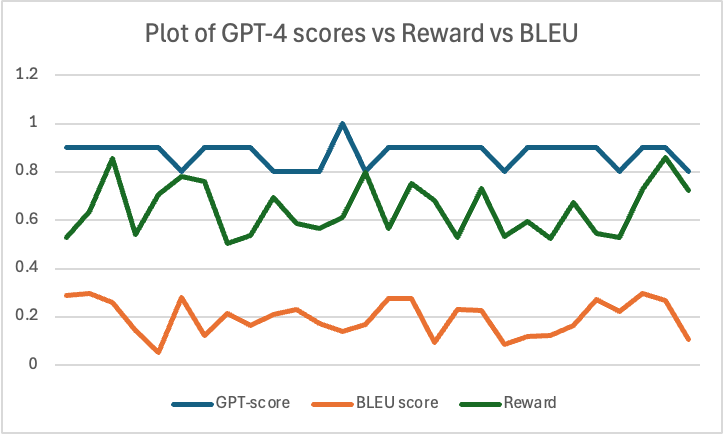}
    \caption{Demonstration of GPT Score \textit{vs.} Reward Score \textit{vs.} BLEU on curated samples having idiomatic expressions, cultural nuances and a mixture of colliqual and formal translations.}
    \label{fig:gpt_scores_vs}
\end{figure}

\noindent\textbf{Mitigation of BLEU score inaccuracies due to contextual matches:}  
The reward model reduces misleading BLEU evaluations caused by idiomatic expressions and cultural nuances that inadvertently match reference translations, leading to false positives. This is further illustrated in the Figure \ref{fig:gpt_scores_vs}, demonstrating that how BLEU and reward compares to GPT based scoring system we can infer that our model's ability to correct such inaccuracies and providing more robust rewarding systems which can also be used to judge translation quality as a metrics.

\noindent\textbf{Reward as \textit{judge}:} Using \textit{reward} as the judge, we calculated the quality of our translations on Fusion-based systems. From Table \ref{tab:BLEUvsReward} we infer that the reward for our approaches SmartGen and SmartGen++ is much higher than LLM-Blender, suggesting that having a training objective where the Ranker and Fusion blocks are inter-dependent results in better final translation quality.

\noindent{\bfseries{Effect of Competitive Correction Block:}}
In both translation directions, we observed that incorporating the CCB Block resulted in improved final translations, as evidenced by the metric scores in Table \ref{tab:perfhin} and \ref{tab:perfeng}, and also increased the reusability of the generated candidates. Our experiment results in Table~\ref{tab:LLM vs improved Rewards} compare the performance of different LLMs as the enhancer LLM $\cG$ in the CCB block (refer to Alg.~\ref{alg:CCB_algo}). We present the average BLEU score for the different LLMs in the two translation directions after utilizing them as $\cG$ in the CCB block. Our findings indicate that the CCB block generates alternative translations that surpass the current translation on BLEU score metrics,


\begin{table}[ht]
\centering
\begin{tabular}{l|c|c}
\hline
\textbf{LLMs} & \textbf{Eng-Hin} & \textbf{Hin-Eng} \\
\hline
LLaMA-3-8B & 28.85 & 42.29 \\
Gajendra & 29.10 & 42.30 \\
Gemma-2-9B & \textbf{29.27} & 42.30 \\
Airavata & 29.25 & 42.28 \\
\hline
\end{tabular}
\caption{Demonstration of LLMs on average enhancement of candidates with CCB block compared to SmartGen performance, 28.92 BLEU (English to Hindi) and 42.73 (Hindi to English)}
\label{tab:LLM vs improved Rewards}
\end{table}

\begin{table}[ht]
\centering
\begin{tabular}{l|c|c}
\hline
\textbf{Methods} & \textbf{BLEU} & \textbf{Reward} \\
\hline

LLM-Blender & 33.01 & 8.59 \\
SmartGen & 33.29 & \textbf{15.35} \\
SmartGen++ & 33.83 & \textbf{15.48} \\

\hline
\end{tabular}
\caption{Comparison of average BLEU and average reward for translation quality judgement on overall test-set.}
\label{tab:BLEUvsReward}
\end{table}



\noindent{\bfseries Note:}
The CCB block, as defined, is independently valuable. While one could use the $Q$-values from the DQN block for correction block admissibility, employing a \textit{separate} reward model allows compatibility with various candidate selection criteria. Additionally, a user-defined reward model enhances flexibility for human alignment and domain adaptation in the final translation.

\section{Conclusion}
In this paper, we identified key limitations in current state-of-the-art ensembling methods for machine translation. We successfully framed the candidate selection problem as a reinforcement learning task, leveraging a Deep Q-Network (DQN) to choose optimal candidates for fusion. This approach reduces inference costs by using selected candidates and improves translation quality through DQN's exploratory capabilities. Additionally, we found that the weakest selected candidate negatively impacts the overall system performance. To address this, we implemented a corrective strategy that enhances translation quality at the cost of increased inference time, which can be of independent interest. Our extensive experiments conducted on benchmark datasets demonstrate that the proposed method yields superior results across various metrics, resulting in SOTA performance.

\section*{Limitations}
One limitation of our work is the fixed value of $K$ (though small), which we plan to make adaptive in future research to improve performance further. Additionally, the criteria for pushing selected candidates to the Competitive Correction Block (CCB) could be refined to enhance both translation quality and time efficiency.

\section*{Ethics Statement}
This work introduces a method for ensembling a pool of NMT systems and LLMs for machine translation tasks. To test our approach we rely on open-source models trained on public data, which may have gender and cultural biases. In our work, we have cited all the open source models and datasets used.

\bibliography{custom}
\newpage
\appendix

\section{Additional details on datasets and methods}
\subsection{Training and Evaluation Data Statistics}
\label{sec:statistics-data}

\begin{table}[!ht]
\small
    \centering
    \begin{tabular}{|c|c|c|} \hline  
         & \textbf{Private Data} & \textbf{Open-Source  Data} \\ \hline  
         \textbf{Training Data} & 98K & - \\ \hline  
         \multirow{4}{*}{\textbf{Test Data}} & - & IN22-Conv-1.5K \\ \cline{2-3}
         & - & IN22Gen-1.02K \\ \cline{2-3}
         & - & WMT-14-2.51K \\ \cline{2-3}
         & 2K &  \\ \cline{2-3}
         & - & Flores-2.01K \\ \hline 
    \end{tabular}
    \caption{Training and Test Data Statistics. The numbers represent pairs of parallel sentences.}
    \label{tab:data_stats}
\end{table}

\subsection{Hyper-parameters used in DQN training}
\begin{table}[ht]
\centering
\begin{tabular}{l|l}
\hline
\textbf{Hyperparameter} & \textbf{Value} \\
\hline
Batch Size & 128 \\
Steps Batch Size & 8 \\
$\gamma$ & 0.99 \\
$\epsilon_{\text{start}}$ & 0.9 \\
$\epsilon_{\text{end}}$ & 0.05 \\
$\epsilon_{\text{decay}}$ & 8000 \\
$\tau$ & $1 \times 10^{-3}$ \\
Target Update & 100 \\
Memory Size & 50000 \\
Learning Rate & $4 \times 10^{-5}$ \\
Episode & 30 \\
Moving Average Window & 100 \\
\hline
\end{tabular}
\caption{Hyperparameters for DQN Training}
\label{tab:dqn-hyperparameters}
\end{table}

\subsection{Prompt used for the CCB block}
\begin{figure*}
    \centering
    \includegraphics[width=\linewidth]{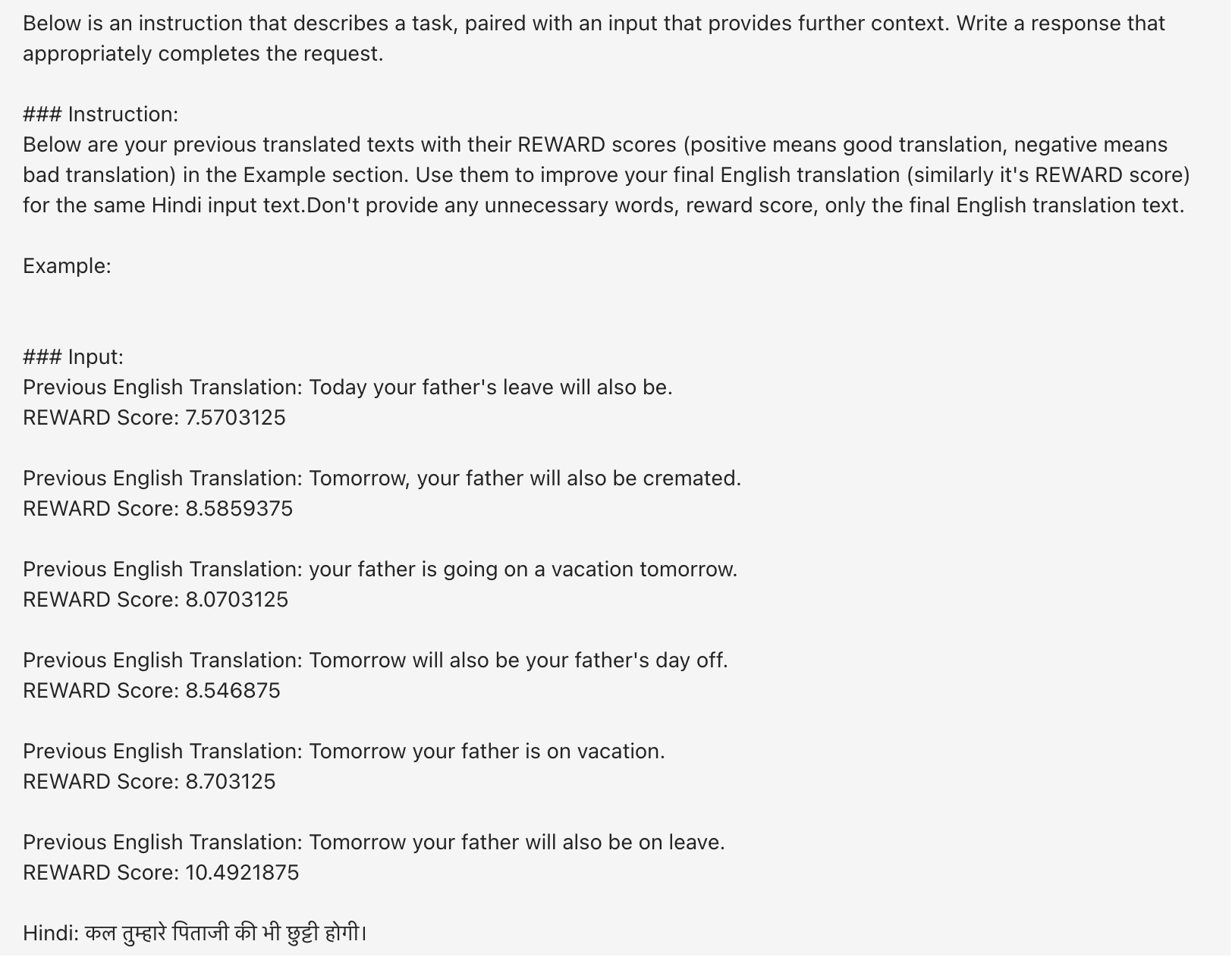}
    \caption{An example of a detailed prompt that we provide to the CCB block LLM $\cG$ along with set of rejected candidate translations and their corresponding scores. }
    \label{fig:ccbprompt}
\end{figure*}

We mention here that we have used ChatGPT for language correction and para-phrasing in drafting the paper.

\end{document}